\documentclass{article}
\usepackage{main}
\usepackage{booktabs}
\usepackage{graphicx}
\graphicspath{{../}}
\usepackage{enumitem}
\usepackage{wrapfig}
\usepackage{natbib}
\usepackage{makecell}
\usepackage{booktabs}
\usepackage{array}
\usepackage{amsmath}
\usepackage{amsfonts}
\usepackage{threeparttable}
\usepackage{multirow}
\usepackage{verbatim}
\usepackage{caption}
\usepackage{longtable}
\usepackage{ragged2e}
\usepackage{array}
\usepackage{mathtools}
\usepackage{bbm,dsfont}
\usepackage{bm}
\usepackage{supertabular}
\usepackage{CJKutf8}
\usepackage{array} 
\usepackage[utf8]{inputenc}
\usepackage[T1]{fontenc}
\usepackage[french,vietnamese,mongolian,greek,english]{babel}
\usepackage{pifont}
\usepackage{xcolor}
\usepackage{tablefootnote}
\usepackage{xspace}
\usepackage{textcomp}
\usepackage{makecell}
\usepackage{lscape} 
\usepackage{siunitx}
\usepackage{listings}
\usepackage{xcolor}
\usepackage{colortbl}
\usepackage[ruled,vlined,linesnumbered]{algorithm2e}
\SetKwComment{tcp}{$\triangleright$~}{}
\definecolor{kuaishoublue}{HTML}{6D9EEB}
\hypersetup{
    colorlinks=true,
}
\definecolor{dt}{gray}{0.7}
\newcommand{\cmark}{\ding{51}}
\newcommand{\xmark}{\ding{55}}
\usepackage{pifont}
\usepackage{bbding}
\usepackage{fontawesome}
\newcolumntype{L}[1]{>{\raggedright\arraybackslash}m{#1}}

\usepackage{scrextend}

\usepackage{tgpagella}
\usepackage{latexsym}
\usepackage[T1]{fontenc}
\usepackage[utf8]{inputenc}
\usepackage{microtype}
\definecolor{mydarkblue}{rgb}{0,0.08,0.45}
\definecolor{citecolor}{HTML}{0071BC}
\usepackage{url}
\usepackage{nicefrac}
\usepackage{changepage}
\usepackage{xargs}
\usepackage{wrapfig,lipsum,booktabs}
\usepackage{longtable}
\usepackage{subcaption}
\usepackage{endnotes}
\usepackage{arydshln}

\usepackage{fancyvrb}
\usepackage{fvextra}
\usepackage{pgfplots}
\usetikzlibrary{pgfplots.groupplots}
\pgfplotsset{compat=1.3}
\usepackage{tikz}
\usetikzlibrary{patterns}

\usepackage[most]{tcolorbox}
\definecolor{blue1}{HTML}{196ab1}
\definecolor{blue2}{HTML}{4886c1}
\definecolor{blue3}{HTML}{5e9bd6}
\definecolor{blue4}{HTML}{77b1e2}
\definecolor{blue5}{HTML}{bdd930}
\definecolor{blue6}{HTML}{dfebf6}

\definecolor{red1}{HTML}{de512c}
\definecolor{red2}{HTML}{f2642d}
\definecolor{red3}{HTML}{f68f58}
\definecolor{red4}{HTML}{febf92}
\definecolor{red5}{HTML}{f8e9c8}

\usepackage[capitalize,noabbrev]{cleveref}
\crefname{section}{Section}{\S\S}
\Crefname{section}{Section}{\S\S}
\crefname{table}{Table}{Tables}
\crefname{figure}{Figure}{Figures}
\crefname{algorithm}{Algorithm}{}
\crefname{equation}{eq.}{}
\crefname{appendix}{Appendix}{}
\crefformat{section}{Section #2#1#3}
\usepackage{multicol}

\DeclareRobustCommand{\ourmethod}{\textsc{Ace-Skill}}

\title{\ourmethod{}: Bootstrapping Multimodal Agents with \\Prioritized and Clustered Evolution}
\author{
\hspace{0pt}Feng Xiong$^{1}$\textsuperscript{$\spadesuit$}, \ 
Zengbin Wang$^{1}$\textsuperscript{$\spadesuit$}, \ 
Yong Wang$^{1}$\textsuperscript{$\clubsuit$}, \ 
Xuecai Hu$^{1}$\textsuperscript{$\clubsuit$}, \ \\
Jinghan He$^{2}$,  \ 
Liang Lin$^{1}$,  \ 
Yuan Liu$^{3}$, \ 
Xiangxiang Chu$^{1}$ \\[6pt]
\hspace{0pt}\mdseries
$^{1}$AMAP, Alibaba Group, \ 
$^{2}$CASIA, \ 
$^{3}$BNU \\[4pt]
}

\newcommand{\eg}{\textit{e.g.}}

\begin{document}
 
\maketitle

{
\renewcommand{\thefootnote}{}
\footnotetext{\textsuperscript{$\spadesuit$} Equal contribution. Work done during the internship at AMAP, Alibaba Group.}
\footnotetext{\textsuperscript{$\clubsuit$} Project leads.}
\addtocounter{footnote}{-2}
}

\begin{abstract}
  Self-evolving agents present a promising path toward continual adaptation by distilling task interactions into reusable knowledge artifacts.
  In practice, this paradigm remains hindered by two coupled bottlenecks: data inefficiency, where costly rollout effort is disproportionately spent on low-value samples rather than informative ones, and knowledge interference, where heterogeneous knowledge stored in shared repositories leads to noisy retrieval and task-misaligned guidance.
  Together, these issues form a self-reinforcing failure loop in which uninformative rollouts yield noisy knowledge, which in turn degrades subsequent rollouts.
  In this work, we introduce \textbf{\ourmethod{}}, a co-evolutionary framework that jointly optimizes rollout allocation and knowledge organization for self-evolving multimodal agents. Specifically, \ourmethod{} combines a \textbf{prioritized sampler} with lazy-decay proficiency tracking to focus rollouts on informative and insufficiently mastered samples, and a \textbf{clustered organizer} that semantically clusters knowledge for cleaner retrieval and more reliable adaptation.
  By improving sampling and organization together, \ourmethod{} turns self-evolution into a virtuous cycle in which more informative rollouts produce higher-quality knowledge that supports stronger subsequent rollouts.
  Across four multimodal tool-use benchmarks, \ourmethod{} delivers strong gains (\eg, +35.46\% relative improvement in Avg@4 accuracy), enabling an open-source 35B MoE model to match or surpass proprietary models. The acquired knowledge also transfers effectively in a zero-shot manner to smaller 9B and 4B models, allowing resource-constrained agents to inherit advanced capabilities without additional training.
  The code has been publicly available at \url{https://github.com/AMAP-ML/Ace-Skill}.
\end{abstract}

\section{Introduction} \label{sec:intro}

    Multimodal Large Language Models (MLLMs) have emerged as a promising foundation for autonomous agents, enabling multimodal reasoning, tool use, and long-horizon decision making in open-ended environments~\citep{bai2025qwen3vl,oai2025o3o4mini,singh2025openai,team2026kimi,Zhipu2026GLM46}. 
    However, strong general-purpose capability does not automatically translate into reliable performance in realistic deployments, where agents must cope with long-tailed tasks, heterogeneous tool interfaces, and continually evolving environments~\citep{li2025tirbench,su2026agentvista,tao2026mmsearchplus,guo2025visualtoolbench}. 
    Adapting the backbone through parameter updates is also an imperfect solution: it is often computationally expensive~\citep{zhou2025agentfly,jiang2026adaptation}, prone to catastrophic forgetting~\citep{zhang2026memrl,jiang2026adaptation}, and poorly suited to real-time adaptation in open-ended settings~\citep{zhou2025agentfly,barnes2025atlas}.

    Against this backdrop, a compelling route toward continual improvement beyond static pre-training is to build self-evolving agents~\citep{xiang2026agentic_self_evolution_survey} that adapt through knowledge bootstrapping~\citep{shinn2023reflexion,zhao2024expel,zhou2025memento,jiang2026xskill,yang2026autoskill,zhang2026memrl,zhang2025memevolve}. 
    With iterative interaction in the environment, these agents externalize what they learn into reusable knowledge artifacts, distilling execution trajectories into experiences and skills that can be reused as context to guide subsequent reasoning and action~\citep{zhao2024expel,xiang2026agentic_self_evolution_survey}. 
    This enables continual adaptation in multimodal settings~\citep{jiang2026xskill}, and in some cases transfer across tasks, domains, or model scales, with little to no modification of the backbone model~\citep{ni2026trace2skill}.

    In practice, realizing this promise remains challenging, especially in realistic multimodal settings with diverse tasks and heterogeneous tools~\citep{li2025tirbench,su2026agentvista,tao2026mmsearchplus,guo2025visualtoolbench}. 
    The core difficulty is to spend limited adaptation budget on informative interactions while keeping the accumulated knowledge reliable across tasks~\citep{zhang2026memrl,wang2026skillx,zheng2026skillrouter,li2026skillsbench}. 
    This exposes two closely coupled structural limitations:
    \begin{itemize}[leftmargin=1.5em, itemsep=1pt, topsep=2pt]
    \item \textbf{Data Inefficiency.} Self-evolving agents depend on costly rollout and reflection loops, whose expense grows in long-horizon multimodal environments with heavy tool use and extended reasoning steps~\citep{allard2026erl,su2026agentvista,zheng2026deeppresenter,du2026mmsearcher}. Yet most pipelines lack a principled way to prioritize under-mastered or high-value samples, wasting scarce adaptation budget on already-solved cases instead of the long tail where improvement matters most.
    \item \textbf{Knowledge Interference.} Shared pool often entangles heterogeneous procedural knowledge, from OCR and spatial reasoning to web grounding and code generation. Without sufficient structure and dependency-aware retrieval, agents may surface noisy or task-misaligned artifacts~\citep{lu2026skill0,tu2026dynamic,chang2026memcollab}, injecting irrelevant guidance, omitting prerequisites, and degrading downstream reasoning and action.
    \end{itemize}

    More importantly, these two bottlenecks are mutually reinforcing, forming a self-perpetuating loop. 
    Misallocated sampling budget oversamples mastered regions while underexploring informative ones, yielding rollouts with limited learning value. 
    Knowledge distilled from such biased trajectories and stored without sufficient organization can then inject noisy or task-misaligned guidance into future interactions, further degrading subsequent rollouts. 
    Breaking this loop requires treating compute allocation and knowledge organization as a coupled optimization problem.

    \begin{figure}[t]
        \centering
        \includegraphics[width=\linewidth]{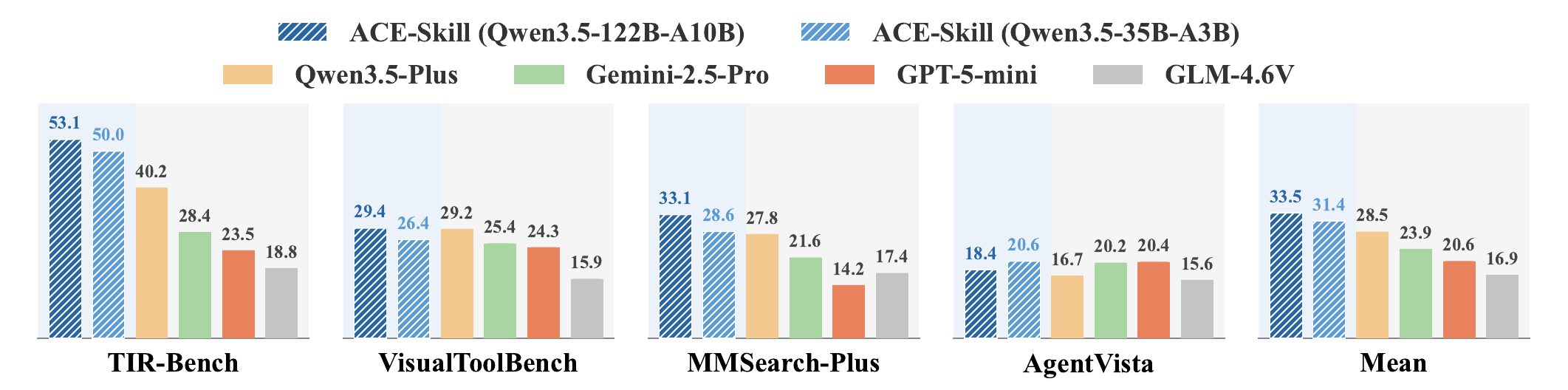}
        \caption{Overall performance of \ourmethod{} across four multimodal tool-use benchmarks~\citep{li2025tirbench,guo2025visualtoolbench,tao2026mmsearchplus,su2026agentvista}. Both \ourmethod{} variants consistently match or surpass prominent closed-source models (\eg, GPT-5-mini~\citep{openai2024gpt5}, Gemini-2.5-Pro~\citep{comanici2025gemini}, Qwen3.5-Plus~\citep{team2026qwen35}) and open-source models (GLM-4.6V~\citep{Zhipu2026GLM46}) under Avg@4 and Pass@4 metrics, despite using significantly smaller backbone models.}
        \label{fig:fig01_performance}
    \end{figure}

    To address these limitations, we introduce \textbf{\ourmethod{}}, a co-evolutionary framework for self-evolving multimodal agents that jointly optimizes compute allocation and knowledge organization. 
    Specifically, \ourmethod{} couples a \textbf{Prioritized Sampler}, which uses lazy-decay proficiency tracking to focus rollout resources on informative and insufficiently mastered samples, with a \textbf{Clustered Organizer}, which semantically clusters knowledge to enable cleaner retrieval and more reliable adaptation.
    By improving sampling and organization together, \ourmethod{} turns self-evolution from a vicious cycle into a virtuous one: more informative rollouts produce higher-quality knowledge, which in turn supports stronger subsequent rollouts. 
    Extensive evaluations across four multimodal tool-use benchmarks show that \ourmethod{} achieves strong gains (\eg, +35.46\% relative improvement in Avg@4 accuracy), enabling an open-source 35B MoE model to match or surpass proprietary models. Furthermore, the acquired knowledge exhibits strong zero-shot transferability to smaller-scale 9B and 4B models.

    To summarize, the main contributions are as follows:
    \begin{itemize}[leftmargin=1.5em, itemsep=1pt, topsep=2pt]
        \item We propose \textbf{\ourmethod{}}, a co-evolutionary framework for self-evolving multimodal agents that jointly optimizes rollout allocation and knowledge organization, addressing the coupled bottlenecks of data inefficiency and knowledge interference in experience and skill bootstrapping.
        \item We introduce two tightly coupled mechanisms: a \textbf{Prioritized Sampler} that leverages lazy-decay proficiency tracking to steer rollout resources toward informative samples, and a \textbf{Clustered Organizer} that structures knowledge into semantically coherent clusters, enabling more dependable adaptation.
        \item Extensive experiments across four multimodal tool-use benchmarks show that \ourmethod{} consistently improves performance, allowing an open-source 35B MoE model to remain competitive with, and in several settings surpass, proprietary baselines. The bootstrapped knowledge also transfers effectively in a zero-shot manner to smaller 9B and 4B models, improving resource-constrained agents without additional training.
    \end{itemize}

\section{\ourmethod{}: Self-evolved Agents with Prioritized and Clustered Evolution} \label{sec:method}
\begin{figure}[t]
      \centering
      \vspace{-0.6cm}
      \includegraphics[width=\linewidth]{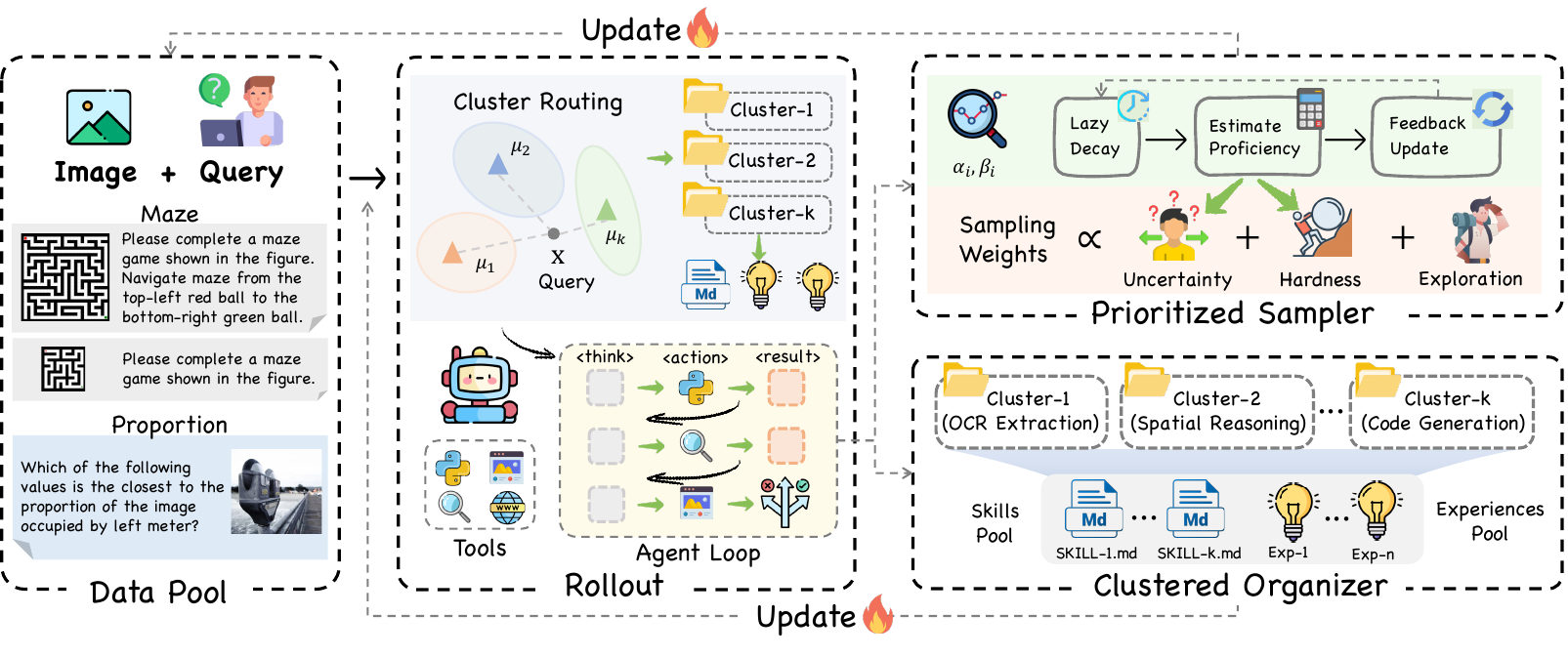} 
      \vspace{-0.7cm}
      \caption{Overview of the \ourmethod{} framework. \ourmethod{} breaks the vicious cycle between data inefficiency and knowledge interference through a prioritized sampler and a clustered organizer.}
      \label{fig:framework}
    \end{figure}

    In this section, we introduce \textbf{\ourmethod{}}, a co-evolutionary framework that bootstraps multimodal agents by jointly optimizing data allocation and knowledge organization, as illustrated in Figure~\ref{fig:framework}. 
    In the following, we first formalize the problem setup and the overall pipeline. Then, we elaborate on the two core components of our framework: a \emph{Prioritized Sampler} that actively selects informative samples for training, and a \emph{Clustered Organizer} that structures and isolates the derived knowledge. 
    A comprehensive step-by-step training procedure is provided in Algorithm~\ref{alg:training}.

\subsection{Framework Overview}
    \noindent\textbf{Problem Setup.}
    Let $\mathcal{D}=\{x_i\}_{i=1}^N$ denote a training pool of $N$ multimodal samples, where each sample $x_i = (I_i, Q_i, A_i)$ comprises images $I_i$, a question $Q_i$, and the ground-truth answer $A_i$.
    To bootstrap the agent's capabilities, the framework employs a \emph{Prioritized Sampler}, parameterized by the proficiency vector $\bm{V}=\{v_i\}_{i=1}^N$, which induces a sampling distribution $p_{\text{sample}}(\cdot \mid \mathcal{D}, \bm{V})$ over $\mathcal{D}$, and a \emph{Clustered Organizer}, denoted by $\{\mathcal{C}_{k}=(\mathcal{E}_k, \mathcal{S}_k)\}_{k=1}^{K}$, where each cluster maintains an experience pool $\mathcal{E}_k$ and a skill $\mathcal{S}_k$. 

    \noindent\textbf{Two-Phase Pipeline.}
    The \ourmethod{} framework operates in two distinct phases:
    \begin{itemize}[leftmargin=*,nosep]
        \item \textbf{Training Phase (\ourmethod{} Bootstrapping):} At each training step $t$, the pipeline evolves the agent through three sequential operations. 
        We first draw a batch $\mathcal{B}_t$ using a biased sampling distribution $p_{\text{sample}}$ derived from the current Bayesian proficiency vector $\bm{V}$ over the dataset $\mathcal{D}$. 
        For each sample $x_i \in \mathcal{B}_t$, let $k_i = c(x_i)$ denote its routed cluster. The agent then performs a rollout on $x_i$, optionally augmented by knowledge retrieved from the routed cluster $\mathcal{C}_{k_i} = (\mathcal{E}_{k_i}, \mathcal{S}_{k_i})$ when available. 
        This yields a scalar reward $r_{t,i} \in [0,1]$ and newly distilled insights $\Delta\mathcal{C}_i = (\Delta\mathcal{E}_i, \Delta\mathcal{S}_i)$. 
        Finally, the reward $r_{t,i}$ is used to update the proficiency estimate $v_i$, while the distilled insights are aggregated back into their routed clusters. Formally:
        \begin{equation}
            \small
            \begin{aligned}
                \mathcal{B}_t &\sim p_{\text{sample}}(\cdot \mid \mathcal{D}, \bm{V}), && \text{\footnotesize\textit{(Prioritized Sampler, \S\ref{sec:sampler})}} \\
                r_{t,i}, \Delta\mathcal{C}_i &= \text{Rollout \& Reflect}(x_i, \mathcal{C}_{k_i}), && \forall x_i \!\in\! \mathcal{B}_t,\; k_i \!=\! c(x_i) \;\; \text{\footnotesize\textit{(Trajectory\ Distillation)}} \\
                v_i &\leftarrow \text{Track}(v_i, r_{t,i}), && \forall x_i \!\in\! \mathcal{B}_t \;\; \text{\footnotesize\textit{(Proficiency Update)}} \\
                \mathcal{C}_k &\leftarrow \text{Integrate}\!\bigl(\mathcal{C}_k, \{\Delta\mathcal{C}_i \mid x_i \!\in\! \mathcal{B}_t,\, k_i \!=\! k\}\bigr), && \forall k \!\in\! \{1,\dots,K\} \;\; \text{\footnotesize\textit{(Cluster Organizer)}}
            \end{aligned}
        \end{equation}
        \vspace{4pt}
        \item \textbf{Inference Phase (Test-Time Injection):} For an unseen sample $x$, the agent routes it to its corresponding cluster $\mathcal{C}_x = (\mathcal{E}_x, \mathcal{S}_x)$. The inference process then sequentially performs knowledge retrieval, task adaptation, and knowledge-guided reasoning:
        \begin{equation}
            \begin{aligned}
                \mathcal{C}_x^{\text{ret}} &= \text{Retrieve}(x, \mathcal{C}_x), && \text{\textit{(Knowledge Retrieval)}} \\
                \mathcal{K}_x &= \text{Adapt}(x, \mathcal{C}_x^{\text{ret}}), && \text{\textit{(Task} \& \textit{Knowledge Adaptation)}} \\
                y &= \text{Agent}(\mathcal{K}_x \oplus x), \quad && \text{\textit{(Knowledge-Guided Reasoning)}}
            \end{aligned}
        \end{equation}
        where $\mathcal{C}_x^{\text{ret}}$ denotes the retrieved relevant tactical experiences and strategic skill, $\mathcal{K}_x$ is the tailored knowledge adapted to the specific sample, and $\oplus$ represents prompt concatenation yielding the final predicted response $y$.
    \end{itemize}

    \begin{algorithm}[t]
        \SetKwComment{tcp}{$\triangleright$~}{}
        \DontPrintSemicolon
        \caption{\ourmethod{} Bootstrapping Procedure}
        \label{alg:training}
        \KwIn{Training pool $\mathcal{D}=\{x_i\}_{i=1}^{N}$; batch size $B$; total steps $T$; number of clusters $K$; sampler parameters $(\rho,\gamma,\epsilon)$}
        \KwOut{Clustered knowledge memory $\mathcal{C}=\{\mathcal{C}_k\!=\!(\mathcal{E}_k, \mathcal{S}_k)\}_{k=1}^K$}
        Partition $\mathcal{D}$ into $K$ semantic clusters and derive the routing rule $c(\cdot)$\;
        Initialize $\{\mathcal{C}_k\!=\!\emptyset\}_{k=1}^K$, $\{\alpha_i\}_{i=1}^{N}\!=\!\{\beta_i\}_{i=1}^{N}\!=\!0$, and $t_i\!=\!0$ for all $x_i$\;
        \For{each step $t = 1$ \KwTo $T$}{
            \For{each $x_i \in \mathcal{D}$}{
                Refresh decayed statistics $\tilde{\alpha}_i \!=\! \alpha_i \cdot \rho^{(t-t_i)}$, $\tilde{\beta}_i \!=\! \beta_i \cdot \rho^{(t-t_i)}$\;
                Estimate proficiency $v_i = (1+\tilde{\alpha}_i)/(2+\tilde{\alpha}_i+\tilde{\beta}_i)$\;
                Set sampling weight $w_t(x_i) \propto \sqrt{v_i(1-v_i)} + \gamma(1-v_i) + \epsilon$\;
            }
            Sample batch $\mathcal{B}_t$ from $\mathcal{D}$ via priorities weights $\{w_t(x_i)\}$  \tcp*[r]{Prioritized Sampler}
            \For{each $x_i \in \mathcal{B}_t$}{
                Route $x_i$ to cluster $k_i \leftarrow c(x_i)$\;
                Perform agent loop on $x_i$ with cluster context $\mathcal{C}_{k_i}\!=\!(\mathcal{E}_{k_i}, \mathcal{S}_{k_i})$\;
                Verify trajectory to obtain reward $r_{t,i}$ and distill $\Delta\mathcal{E}_i, \Delta\mathcal{S}_i$ \tcp*[r]{Traj. Distillation}
                Update proficiency statistics $\alpha_i \!\leftarrow\! \tilde{\alpha}_i + r_{t,i}$, $\beta_i \!\leftarrow\! \tilde{\beta}_i + (1\!-\!r_{t,i})$, $t_i \!\leftarrow\! t$ \tcp*[r]{Proficiency Update}
            }
            \For{each cluster $k = 1$ \KwTo $K$}{
                Update $\mathcal{E}_k$ using $\{\Delta\mathcal{E}_i \mid x_i \in \mathcal{B}_t,\; k_i = k\}$ via similarity-aware retrieve-and-merge\;
                Update $\mathcal{S}_k$ using $\{\Delta\mathcal{S}_i \mid x_i \in \mathcal{B}_t,\; k_i = k\}$ via merge-and-compress \tcp*[r]{Clustered Organizer}
            }
        }
    \end{algorithm}

\subsection{Prioritized Sampler} \label{sec:sampler}
    Most existing bootstrapping pipelines still rely on sequential or uniform sample traversal~\citep{jiang2026xskill,zhou2026memento}, allocating rollout budget to already-solved cases rather than informative, under-mastered ones.
    To maximize data utility, we introduce the \emph{Prioritized Sampler}, which concentrates resources on the most informative samples. It maintains an evolving per-sample proficiency estimate $v_i$, converts it into a sampling distribution, and updates $v_i$ using rollout outcomes from each round.
    
    \noindent\textbf{Tracking Proficiency.}
    As bootstrapping proceeds, the agent's capability is inherently non-stationary: a sample that is difficult early on may later become routine. To model this evolution, we maintain for each sample $x_i$ a Beta posterior $\text{Beta}(1+\tilde{\alpha}_i,\;1+\tilde{\beta}_i)$ over its success probability $v_i$, with $(\alpha_i,\beta_i)$ initialized to $(0.5,0.5)$.
    To keep $v_i$ aligned with the agent's current capability rather than outdated observations, we adopt lazy decay~\citep{xu2025singlestream}, which discounts the statistics of $x_i$ only when it is revisited, using a factor $\rho$ raised to the elapsed steps since its last update $t_i$:
    \begin{equation}\label{eq:lazy-decay}
        \tilde{\alpha}_i = \alpha_i \cdot \rho^{(t - t_i)}, \qquad
        \tilde{\beta}_i = \beta_i \cdot \rho^{(t - t_i)}, \qquad
        v_i = \frac{1 + \tilde{\alpha}_i}{2 + \tilde{\alpha}_i + \tilde{\beta}_i}.
    \end{equation}

    \noindent\textbf{Prioritized Sampling.}
    Given the up-to-date $v_i$, we convert proficiency estimates into a sampling distribution that prioritizes the most informative samples. Each sample is assigned a sampling weight balancing three desiderata:
    \begin{equation}\label{eq:weight}
        w_t(x_i) \propto \underbrace{\sqrt{v_i(1-v_i)}}_{\text{uncertainty}} + \underbrace{\gamma\,(1-v_i)}_{\text{hardness}} + \underbrace{\epsilon}_{\text{exploration}}.
    \end{equation}
    The uncertainty term peaks near $v_i \!\approx\! 0.5$, where uncertainty is greatest; the hardness term ($\gamma > 0$) biases sampling toward persistently difficult samples; and $\epsilon$ provides an exploration floor that prevents starvation and preserves broad coverage. We then draw a batch $\mathcal{B}_t$ of size $B$ from $\mathcal{D}$ according to the resulting weights $\{w_t(x_i)\}$.

    \vspace{2pt}
    \noindent\textbf{Updating Proficiency Estimates.}
    After rollout, the observed reward $r_{t,i} \in [0,1]$ is incorporated into the decayed statistics to update the proficiency estimate:
    \begin{equation}\label{eq:beta-update}
        \alpha_i \leftarrow \tilde{\alpha}_i + r_{t,i}, \qquad
        \beta_i \leftarrow \tilde{\beta}_i + (1 - r_{t,i}), \qquad
        t_i \leftarrow t.
    \end{equation}
    The resulting $v_i$ then determines the sampling priorities for the next round, closing the proficiency-tracking and prioritized-sampling loop.

    \begin{table}[t]
        \centering
        \caption{Dual-granularity design within each cluster $\mathcal{C}_k = (\mathcal{E}_k, \mathcal{S}_k)$.}
        \label{tab:dual-granularity}
        \renewcommand{\arraystretch}{1.15}
        \setlength{\tabcolsep}{4pt}
        \resizebox{\textwidth}{!}{
        \begin{tabular}{lll}
            \toprule
            & \textbf{Tactical Experience Cluster} $\mathcal{E}_k$ & \textbf{Strategic Skill Cluster} $\mathcal{S}_k$ \\
            \midrule
            \textbf{Granularity} & Instance-level & Category-level \\
            \textbf{Content} & Concise textual lessons (\eg, tool tips, error fixes) & Structured SOP (workflows, heuristics, pitfalls) \\
            \textbf{Format} & Key-value entries (embedding-indexed) & Markdown document \\
            \textbf{Retrieval} & Top-$k$ cosine similarity & Full document, adapted per task \\
            \textbf{Lifecycle} & Critique $\to$ Embed-gate merge $\to$ Reduce to $L$ & Generate $\to$ Merge $\to$ Refine to $W$ words \\
            \textbf{Capacity bound} & $|\mathcal{E}_k| \leq L$ entries & $|\mathcal{S}_k| \leq W$ words \\
            \textbf{Best suited for} & Samples closely matching a prior instance & Novel samples within the same category \\
            \bottomrule
        \end{tabular}
        }
    \end{table}

\subsection{Clustered Organizer} \label{sec:memory}
    A monolithic knowledge bank~\citep{jiang2026xskill,tu2026dynamic} pools knowledge from heterogeneous tasks (\eg, OCR, spatial reasoning, and code generation) in a single repository, resulting in cross-type retrieval noise and unbounded growth. To address this issue, the \emph{Clustered Organizer} routes samples into $K$ semantic clusters. Each cluster $\mathcal{C}_k = (\mathcal{E}_k, \mathcal{S}_k)$ maintains a tactical experience pool $\mathcal{E}_k$ and a strategic skill document $\mathcal{S}_k$, allowing knowledge to be organized and evolved at complementary granularities.

    \noindent\textbf{Semantic Clustering and Routing.}\label{sec:cluster-route}
    Since task categories (\eg, OCR, spatial reasoning, and code generation) are induced primarily by input semantics and remain comparatively stable during training, the cluster structure can be specified \textit{a priori} rather than adapted online. This design avoids the optimization instability and auxiliary hyperparameter sensitivity introduced by dynamic clustering, while preserving a unified routing rule across both training and inference.
    Formally, let $f_\theta(x)$ denote the text embedding of the description associated with sample $x$. We apply $K$-means~\citep{likas2003kmeans} to these embeddings to obtain centroids $\{\mu_k\}_{k=1}^K$, and assign each sample $x$ to its nearest centroid:
    \begin{equation}
        c(x) = \arg\min_{1 \leq k \leq K} \left\| f_\theta(x) - \mu_k \right\|_2^2.
    \end{equation}
    This deterministic assignment confines all subsequent retrieval, storage, and update operations to a single semantically coherent cluster, thereby mitigating interference arising from heterogeneous task types.

    \noindent\textbf{Evolving Tactical Experience Pool $\mathcal{E}_k$.}\label{sec:experience}
    Within each cluster, $\mathcal{E}_k$ maintains instance-level lessons that encode fine-grained failure modes and corrective actions for closely related samples. Given a new experience candidate $\Delta\mathcal{E}_i$, we first retrieve all existing entries whose similarity to $\Delta\mathcal{E}_i$ exceeds a threshold $\tau$, yielding a neighbor set $\mathcal{N} = \{e \in \mathcal{E}_k \mid \text{sim}(\Delta\mathcal{E}_i, e) > \tau\}$. The pool is then updated as follows:
    \begin{equation}
        \mathcal{E}_k \leftarrow
        \begin{cases} 
            \mathcal{E}_k \cup \{\Delta\mathcal{E}_i\}, & \text{if } \mathcal{N} = \emptyset \\
            (\mathcal{E}_k \setminus \mathcal{N}) \cup \{\text{Merge}(\Delta\mathcal{E}_i, \mathcal{N})\}, & \text{otherwise}
        \end{cases}
    \end{equation}
    where $\tau$ denotes the similarity threshold. If no sufficiently similar entry is found, the candidate is inserted directly; otherwise, $\Delta\mathcal{E}_i$ and its retrieved neighbors are consolidated into a single entry via LLM-based merging. To enforce the capacity constraint $|\mathcal{E}_k| \leq L$, we further apply a compaction step that iteratively merges the most similar pair until the pool size falls within the budget.

    \noindent\textbf{Evolving Strategic Skill Pool $\mathcal{S}_k$.}\label{sec:skill}
    Whereas $\mathcal{E}_k$ captures instance-specific lessons, novel samples may not admit close historical analogues. To complement such fine-grained experience, $\mathcal{S}_k$ distills recurrent, category-level procedures into a single structured document. After each rollout, the LLM extracts a skill update $\Delta\mathcal{S}_i$, which is integrated into the current skill document and compressed to satisfy a budget $W$:
    \begin{equation}
        \mathcal{S}_k \leftarrow \text{Compress}\Big(\text{Merge}\big(\mathcal{S}_k, \Delta\mathcal{S}_i\big), \; W\Big).
    \end{equation}
    As systematically compared in Table~\ref{tab:dual-granularity}, the cluster-isolated pair $(\mathcal{E}_k, \mathcal{S}_k)$ provides knowledge at two complementary levels of granularity: $\mathcal{E}_k$ offers fine-grained tactical guidance for samples resembling previously encountered cases, while $\mathcal{S}_k$ supplies higher-level strategic guidance for less familiar or unseen instances. Together, they complete the co-evolutionary loop with the sampler (\S\ref{sec:sampler}).

\section{Experiments} \label{sec:experi}
    \subsection{Experimental Settings}
    \noindent\textbf{Datasets and Benchmarks.} 
    To comprehensively evaluate multimodal tool use and agent behavior, we conduct experiments across four diverse benchmarks. These include one code-based benchmark and three general tool-use benchmarks as follows. If not specific, we follow the recent metric setting~\citep{jiang2026xskill} and report \texttt{Avg@4} and \texttt{Pass@4} across all evaluated datasets.

    \begin{itemize}[leftmargin=*, itemsep=1pt, topsep=2pt]
      \item \textbf{TIR-Bench}~\citep{li2025tirbench} is a code-based benchmark designed for Tool-Integrated Reasoning. It evaluates the agent's ability to solve complex tasks by writing and executing code in interactive environments, covering domains such as mathematics, data science, and general programming.
      \item \textbf{VisualToolBench}~\citep{guo2025visualtoolbench} is a general tool-use benchmark focusing on selecting and executing tools grounded in visual context. It covers a broad spectrum of vision-language tool invocations, ranging from image editing and visual generation to complex visual comprehension tasks.
      \item \textbf{MMSearch-Plus}~\citep{tao2026mmsearchplus} is a general tool-use benchmark tailored for handling complex multimodal search queries. It evaluates the capacity to retrieve, analyze, and synthesize information from the web, spanning domains like general knowledge, real-time news, and specialized online information seeking.
      \item \textbf{AgentVista}~\citep{su2026agentvista} is a general tool-use benchmark for executing visually grounded, long-horizon tasks. It assesses sequential decision-making in dynamic environments, covering practical operating system navigation, graphical user interface (GUI) operations, and daily web applications.
    \end{itemize}
    
    \noindent\textbf{Tool Library.}
    Following the recent XSkill~\citep{jiang2026xskill}, we empower our agent with four types of tools to handle various tasks as follows.
    \begin{itemize}[leftmargin=*, itemsep=1pt, topsep=2pt]
      \item \textbf{Code Interpreter}: A stateful Jupyter kernel~\citep{kluyvertextordfeminine12016jupyter} that executes Python code for algorithmic computation, data manipulation, and image processing, supporting persistent execution and pre-installed with libraries such as NumPy, Pandas, and OpenCV.
      \item \textbf{Image Search}: A dual-mode search utility powered by the Google Images Search API~\citep{google_search} that supports both text-to-image queries and reverse image search, which utilizes ImgBB~\citep{imgbb} for uploading local images prior to the Google search.
      \item \textbf{Web Search}: A search engine interface using the Google Search API~\citep{google_search} to retrieve up-to-date information, returning relevant titles, URLs, and text snippets.
      \item \textbf{Visit}: A webpage content extraction tool that utilizes the open-source Crawl4ai~\citep{crawl4ai2024}, along with Scrape Serper~\citep{google_search} as a fallback mechanism, to parse the main textual content from specific URLs, featuring an optional capability to generate goal-oriented summaries using LLMs.
    \end{itemize}

    \noindent\textbf{Baselines and Backbones.}
    To assess the effectiveness of \ourmethod{}, we adopt Qwen3.5-35B-A3B~\citep{team2026qwen35} as the primary backbone and compare against two groups of recent models: (1)~\emph{Proprietary models}: o4-mini~\citep{oai2025o3o4mini}, GPT-5-mini~\citep{openai2024gpt5}, Gemini-2.5-Pro~\citep{comanici2025gemini}, and Qwen3.5-Plus~\citep{team2026qwen35}; (2)~\emph{Open-source models}: GLM-4.6V~\citep{Zhipu2026GLM46} and Kimi-K2.5~\citep{team2026kimi}.
    For each model, we report two settings: the model's \emph{direct answer} and \emph{vanilla tool calling}, which equips the model with our standardized tool library described above.
    Additionally, we extend \ourmethod{} to the larger Qwen3.5-122B-A10B and the smaller Qwen3.5-9B/4B to enable scalability and transfer experiments, thereby broadening its applicability across diverse deployment scenarios.

    \noindent\textbf{Implementation Details.}
    \ourmethod{} conducts end-to-end self-evolution: the same MLLM serves as the reasoning agent, trajectory summarizer, experience \& skill merger, and verifier, all operating in no-thinking mode. 
    Training runs for 50 steps with batch size 8, where each sample undergoes 1 rollout of up to 10 agentic turns for TIR-Bench and 20 for other benchmarks, at temperature $0.7$.
    For the prioritized sampler, we set temporal decay $\rho\!=\!0.95$, difficulty bias $\gamma\!=\!0.4$, and base exploration $\epsilon\!=\!0.1$. 
    Text embeddings for clustering and retrieval are computed by \texttt{text-embedding-3-small}~\citep{openai2024embeddings}, with a cosine similarity threshold $\tau\!=\!0.70$ for experience merging. 
    Each cluster's experience pool is capped at $L\!=\!120$ entries, and the skill document is refined when exceeding $W\!>\!1{,}000$ words.
    Additionally, for models smaller than 9B, we adopt Qwen3.5-9B as the verifier, as sub-9B models exhibit hallucinated judgments when serving as a verifier.
    For computational resources, Qwen3.5-Plus and Kimi K2.5 are accessed via official APIs, while GLM-4.6V and our adopted 35B/122B models are deployed locally with the SGLang framework~\citep{zheng2024sglang} on 8$\times$H20 (96 GB) GPUs.

\subsection{Main Results}

    \begin{table}[t]
        \centering
        \setlength{\tabcolsep}{2mm}
        \renewcommand\arraystretch{1.3}
        \caption{Main results (Avg@4 and Pass@4) across four multimodal tool-use benchmarks. Each model is evaluated under its native direct answer and with our vanilla tool library. ``$^*$'' indicates that the results are sourced from~\citep{jiang2026xskill}, and the others are reproduced using official checkpoints or APIs. \textbf{Bold} indicates the best result.}
        \label{tab:main}
        \resizebox{\textwidth}{!}{
        \begin{tabular}{lcccccccccc}
        \toprule
        \multicolumn{1}{c}{\multirow{2}{*}{\textbf{Model}}} & \multicolumn{2}{c}{\textbf{TIR-Bench}} & \multicolumn{2}{c}{\textbf{VisualToolBench}} & \multicolumn{2}{c}{\textbf{MMSearch-Plus}} & \multicolumn{2}{c}{\textbf{AgentVista}} & \multicolumn{2}{c}{\textbf{Mean}} \\ 
        \cmidrule{2-11} 
        \multicolumn{1}{c}{} & Avg@4 & \multicolumn{1}{c}{Pass@4} & Avg@4 & \multicolumn{1}{c}{Pass@4} & Avg@4 & \multicolumn{1}{c}{Pass@4} & Avg@4 & \multicolumn{1}{c}{Pass@4} & Avg@4 & \multicolumn{1}{c}{Pass@4} \\ 
        \midrule
        \multicolumn{11}{c}{\emph{\textbf{{Proprietary Multimodal Models}}}} \\
        \midrule
        o4-mini$^*$~\citep{oai2025o3o4mini} & 20.13 & 45.00 & 14.72 & 27.57 & 5.92 & 12.32 & 19.04 & 25.69 & 14.95 & 27.65 \\
        \quad $\hookrightarrow$ Vanilla Tools & 24.62 & 49.50 & 19.63 & 34.11 & 15.88 & 21.80 & 18.12 & 29.36 & 19.56 & 33.69 \\
        \hdashline
        GPT-5-mini$^*$~\citep{openai2024gpt5} & 20.00 & 46.50 & 13.90 & 22.90 & 3.08 & 6.64 & 18.58 & 28.44 & 13.89 & 26.12 \\
        \quad $\hookrightarrow$ Vanilla Tools & 23.50 & 50.50 & 24.30 & 37.85 & 14.22 & 20.38 & 20.41 & 35.78 & 20.61 & 36.13 \\
        \hdashline
        Gemini-2.5-Pro$^*$~\citep{comanici2025gemini} & 21.37 & 40.00 & 20.91 & 28.97 & 10.43 & 19.43 & 17.89 & 28.44 & 17.65 & 29.21 \\
        \quad $\hookrightarrow$ Vanilla Tools & 28.38 & 54.00 & 25.35 & 40.65 & 21.56 & 35.55 & 20.18 & 33.94 & 23.87 & 41.04 \\
        \hdashline
        \multicolumn{1}{l}{Qwen3.5-Plus~\citep{team2026qwen35}} & 29.88 & 47.50 & 17.17 & 28.04 & 6.64 & 10.43 & 11.47 & \multicolumn{1}{c}{18.35} & 16.29 & 26.08 \\
        \multicolumn{1}{l}{\quad $\hookrightarrow$ Vanilla Tools} & 40.25 & 62.50 & 29.21 & 49.07 & 27.84 & 46.45 & 16.74 & 31.19 & 28.51 & 47.30 \\
        \midrule
        \multicolumn{11}{c}{\emph{\textbf{{Open-Source Multimodal Models}}}} \\
        \midrule
        GLM-4.6V~\citep{Zhipu2026GLM46} & 17.50 & 28.00 & 17.76 & 34.11 & 3.67 & 8.06 & 15.14 & 24.77 & 13.52 & 23.74 \\
        \multicolumn{1}{l}{\quad $\hookrightarrow$ Vanilla Tools} & 18.75 & 42.00 & 15.89 & 31.78 & 17.42 & 34.60 & 15.60 & 31.19 & 16.92 & 34.89 \\
        \hdashline
        Kimi-K2.5~\citep{team2026kimi} & 17.37 & 28.00 & 12.85 & 19.63 & 10.78 & 18.48 & 18.12 & 30.28 & 14.78 & 24.10 \\
        \multicolumn{1}{l}{\quad $\hookrightarrow$ Vanilla Tools} & 19.38 & 37.00 & 21.03 & 34.11 & 33.53 & 48.34 & 18.35 & 31.19 & 23.07 & 37.66  \\
        \midrule
        \multicolumn{11}{c}{\emph{\textbf{{Baseline and \ourmethod{}}}}} \\
        \midrule
        Qwen3.5-35B-A3B~\citep{team2026qwen35} & 23.00 & 43.00 & 15.65 & 28.04 & 3.67 & 7.58 & 10.32 & 17.43 & 13.16 & 24.01 \\
        \quad $\hookrightarrow$ Vanilla Tools & 39.37 & 65.00 & 21.38 & 39.72 & 16.59 & 28.44 & 15.37 & 32.11 & 23.18 & 41.32  \\
        \quad $\hookrightarrow$ \textit{w/} XSkill~\citep{jiang2026xskill} & 40.62 & 68.00 & 22.43 & 35.51 & 23.93 & 41.23 & 16.51 & 29.36 & 25.87 & 43.53 \\
        \rowcolor{black!10} \quad \textbf{$\hookrightarrow$ \textit{w/} \ourmethod{}} & \textbf{50.00} & \textbf{71.50} & \textbf{26.40} & \textbf{41.59} & \textbf{28.55} & \textbf{43.13} & \textbf{20.64} & \textbf{36.70} & \textbf{31.40} & \textbf{48.23} \\
        \bottomrule
        \end{tabular}
        }
     \end{table}
     
\begin{table}[t]
      \centering
      \setlength{\tabcolsep}{2mm}
      \renewcommand\arraystretch{1.3}
      \caption{Scalability and zero-shot transfer results (Avg@4 and Pass@4). Top: scaling \ourmethod{} to Qwen3.5-122B-A10B. Bottom: directly injecting knowledge bootstrapped by 35B-A3B into smaller models (9B, 4B) without retraining. \textbf{Bold} indicates the best result.}
      \label{tab:scale_transfer}
      \resizebox{\textwidth}{!}{
          \begin{tabular}{lcccccccccc}
          \toprule
          \multicolumn{1}{c}{\multirow{2}{*}{\textbf{Model}}} & \multicolumn{2}{c}{\textbf{TIR-Bench}} & \multicolumn{2}{c}{\textbf{VisualToolBench}} & \multicolumn{2}{c}{\textbf{MMSearch-Plus}} & \multicolumn{2}{c}{\textbf{AgentVista}} & \multicolumn{2}{c}{\textbf{Mean}} \\ 
          \cmidrule{2-11} 
          \multicolumn{1}{c}{} & Avg@4 & \multicolumn{1}{c}{Pass@4} & Avg@4 & \multicolumn{1}{c}{Pass@4} & Avg@4 & \multicolumn{1}{c}{Pass@4} & Avg@4 & \multicolumn{1}{c}{Pass@4} & Avg@4 & \multicolumn{1}{c}{Pass@4} \\ 
          \midrule
          \multicolumn{11}{c}{{\emph{\textbf{Exploring More Scalable Model (122B-A10B)}}}} \\
          \midrule
          Qwen3.5-122B-A10B~\citep{team2026qwen35} & 22.13 & 41.00 & 13.90 & 22.90 & 4.03 & 9.48 & 9.86 & 18.35 & 12.48 & 22.93 \\
          \quad $\hookrightarrow$ Vanilla Tools & 46.12 & 74.50 & 23.83 & 40.19 & 29.23 & 45.97 & 15.60 & 31.19 & 28.70 & 47.96 \\
          \rowcolor{black!10} \quad \textbf{$\hookrightarrow$ \textit{w/} \ourmethod{}} & \textbf{53.12} & \textbf{75.00} & \textbf{29.44} & \textbf{50.47} & \textbf{33.06} & \textbf{49.29} & \textbf{18.35} & \textbf{33.03} & \textbf{33.49} & \textbf{51.95} \\
          \midrule
          \multicolumn{11}{c}{{\emph{\textbf{Transfer Evaluations of Experiences and Skills (35B-A3B $\xrightarrow{}$ 9B, 4B)}}}} \\
          \midrule
          \multicolumn{1}{l}{Qwen3.5-9B~\citep{team2026qwen35}} & 14.75 & 28.50 & 10.16 & 21.03 & 1.66 & 4.27 & 4.59 & 11.93 & 7.79 & 16.43 \\
          \quad $\hookrightarrow$ Vanilla Tools & 26.75 & 50.50 & 18.11 & 34.11 & 14.45 & 27.49 & 9.17 & 22.94 & 17.12 & 33.76 \\
          \rowcolor{black!10} \quad \textbf{$\hookrightarrow$ \textit{w/} \ourmethod{}} & \textbf{39.62}  & \textbf{66.00} & \textbf{20.56} & \textbf{35.51} & \textbf{24.53} & \textbf{37.91} & \textbf{17.43} & \textbf{33.03} & \textbf{25.54} & \textbf{43.11} \\
          \hdashline
          \multicolumn{1}{l}{Qwen3.5-4B~\citep{team2026qwen35}} & 10.25 & 28.50 & 5.96 & 14.49 & 0.59 & 1.42 & 4.59 & 12.84 & 5.35 & 14.31 \\
          \quad $\hookrightarrow$ Vanilla Tools & 25.12 & 46.00 & 12.85 & \textbf{30.37} & 14.10 & 26.54 & 8.03 & 22.02 & 15.03 & 31.23 \\
          \rowcolor{black!10} \quad \textbf{$\hookrightarrow$ \textit{w/} \ourmethod{}} & \textbf{33.50} & \textbf{61.50} & \textbf{15.07} & 29.91 & \textbf{19.31} & \textbf{29.38} & \textbf{11.93} & \textbf{22.94} & \textbf{19.95} & \textbf{35.93} \\
          \bottomrule
          \end{tabular}
    }
   \vspace{-0.3cm}
   \end{table}
    
    \noindent\textbf{Overall Performance.}
    Table~\ref{tab:main} summarizes results across four benchmarks. With Qwen3.5-35B-A3B as the backbone, \ourmethod{} achieves a mean Avg@4 of 31.40 and Pass@4 of 48.23, improving over the Vanilla Tools baseline by +8.22 and +6.91 points, respectively. 
    The gains are consistent across all benchmarks, with the largest improvement on TIR-Bench (+10.63 Avg@4) and MMSearch-Plus (+11.96 Avg@4), demonstrating that bootstrapped experiences and skills are particularly beneficial for code-intensive and search-intensive tasks.
    Compared with proprietary models, \ourmethod{} with an open-source 35B backbone surpasses GPT-5-mini (+10.79 Avg@4), Gemini-2.5-Pro (+7.53 Avg@4), and even the Qwen3.5-Plus API (+2.89 Avg@4) under the Vanilla Tools setting. 
    This indicates that self-evolved knowledge can close the gap between open-source and proprietary models without model weight updates.

    \noindent\textbf{Scaling to Larger Backbone.}
    As shown at the top of Table~\ref{tab:scale_transfer}, \ourmethod{} continues to produce consistent gains when applied to the larger Qwen3.5-122B-A10B, reaching a mean Avg@4 of 33.49 and Pass@4 of 51.95, the highest among all evaluated configurations. 
    Although the relative improvement over Vanilla Tools (+4.79 Avg@4) is smaller than in the 35B setting, which aligns with the intuition that stronger base models already handle more tasks correctly, leaving less room for knowledge augmentation. 
    Nevertheless, the gains remain positive and stable across all four benchmarks, confirming that \ourmethod{} does not saturate and can reliably benefit models of increasing capacity.

    \noindent\textbf{Zero-Shot Transfer Across Model Scales.}
    A key practical question is whether the experiences and skills accumulated by a large model can benefit smaller ones without retraining. We directly inject the knowledge bootstrapped by Qwen3.5-35B-A3B into the smaller Qwen3.5-9B and Qwen3.5-4B at inference time. 
    As shown in the bottom of Table~\ref{tab:scale_transfer}, the transferred knowledge yields substantial gains: on Qwen3.5-9B, \ourmethod{} improves TIR-Bench from 26.75 to 39.62 Avg@4 (+12.87) and from 50.50 to 66.00 Pass@4 (+15.50), even approaching the 35B Vanilla Tools baseline (39.37 Avg@4). 
    Notably, the gains persist on the smaller Qwen3.5-4B, demonstrating that structured, cluster-isolated knowledge remains effective even for resource-constrained models with limited reasoning capacity.
    
    \noindent\textbf{Cross-Bench Generalization (OOD).}
    Beyond in-domain improvements, a crucial question is whether the bootstrapped knowledge can transfer across domain boundaries. Following the OOD evaluation protocol of XSkill~\citep{jiang2026xskill}, we conduct two OOD transfer experiments, as summarized in Table~\ref{tab:OOD}. We inject experiences and skills bootstrapped \emph{exclusively} on VisualToolBench into TIR-Bench, and those from MMSearch-Plus into AgentVista.
    On Qwen3.5-35B-A3B, OOD transfer lifts TIR-Bench Avg@4 from 39.37 to 45.50 (+6.13) and AgentVista from 15.37 to 18.58 (+3.21). The gains persist on the larger Qwen3.5-122B-A10B (+4.13 on TIR-Bench and +4.35 on AgentVista), indicating that the bootstrapped knowledge encodes transferable reasoning strategies rather than benchmark-specific shortcuts.
    This cross-domain transferability can be attributed to \ourmethod{}'s cluster-isolated organization: semantically coherent clusters distill general-purpose skills (\eg, structured code execution workflows, iterative search-and-verify patterns) that remain applicable even when the target task format differs substantially from the source.

\begin{figure}[t]
        \centering
        \includegraphics[width=\textwidth]{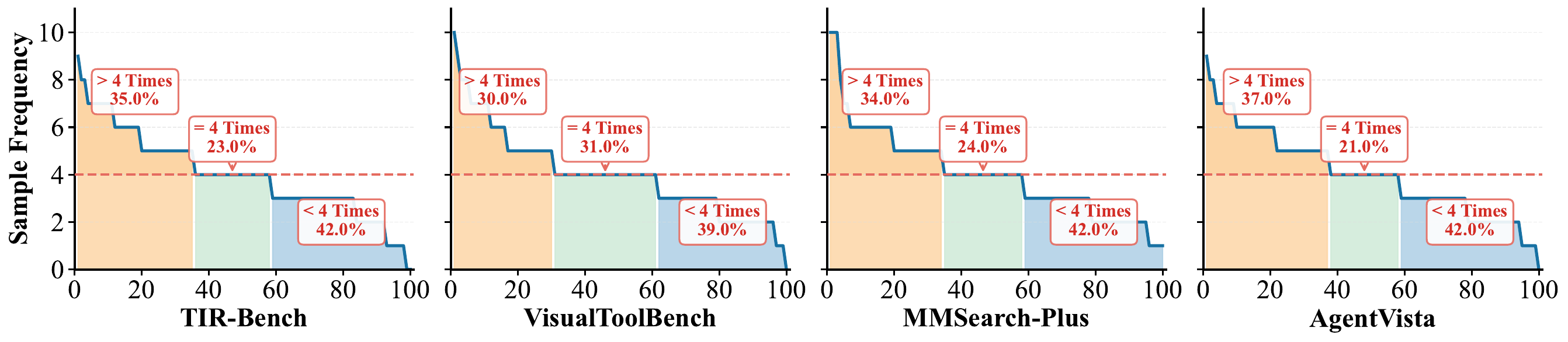}
        \caption{Per-sample selection frequency of the Prioritized Evolution Sampler over 4 epochs on four benchmarks. Under uniform sampling, every sample would be visited exactly 4 times.}
        \label{fig:sampling_analysis}
    \end{figure}

    \noindent\textbf{Sampling Distribution Analysis.}
    Figure~\ref{fig:sampling_analysis} shows the per-sample selection frequency over 4 epochs. Under uniform sampling, every sample would be visited exactly 4 times. Our Prioritized Evolution Sampler deviates substantially: across all benchmarks, over 30\% of samples are visited more than 4 times, as they remain uncertain or challenging (low $V_i$) and receive higher selection weight $w_t$ (Eq.~\ref{eq:weight}). Conversely, over 39\% of samples are visited fewer than 4 times, as the agent has already mastered them ($V_i \!\approx\! 1$) and the sampler automatically reduces their priority. This shows that the sampler successfully reallocates compute from easy samples to the ones that matter most.

    \begin{table}[]
        \centering
        \setlength{\tabcolsep}{6mm} 
        \renewcommand\arraystretch{1.25}
        \caption{Cross-domain generalization (OOD) performance (Avg@4). Experiences and skills are bootstrapped \emph{only} on the source benchmark and directly transferred to a held-out target benchmark \emph{without} any additional training or adaptation on the target domain. $\bigtriangleup$ denotes the absolute improvement over Vanilla Tools.}
        \label{tab:OOD}
        \resizebox{\textwidth}{!}{
            \begin{tabular}{lcccc}
                \toprule
                Model & \multicolumn{2}{c}{VisualToolBench $\rightarrow$ TIR-Bench} & \multicolumn{2}{c}{MMSearch-Plus $\rightarrow$ AgentVista} \\
                \cmidrule(lr){2-3} \cmidrule(lr){4-5}
                & Avg@4 & $\bigtriangleup$ & Avg@4 & $\bigtriangleup$ \\ 
                \midrule
                Qwen3.5-35B-A3B~\citep{team2026qwen35} & 39.37 & - & 15.37 & - \\
                \rowcolor{black!10} \quad $\hookrightarrow$ \emph{+ OOD Transfer} & 45.50 & +6.13 & 18.58 & +3.21 \\ 
                \hdashline
                Qwen3.5-122B-A10B~\citep{team2026qwen35} & 46.12 & - & 15.60 & - \\
                \rowcolor{black!10} \quad $\hookrightarrow$ \emph{+ OOD Transfer} & 50.25 & +4.13 & 19.95 & +4.35 \\ 
                \bottomrule
            \end{tabular}
        }
    \end{table}

    \subsection{Ablation Study}
\begin{figure}[t]
        \centering
        \includegraphics[width=\linewidth]{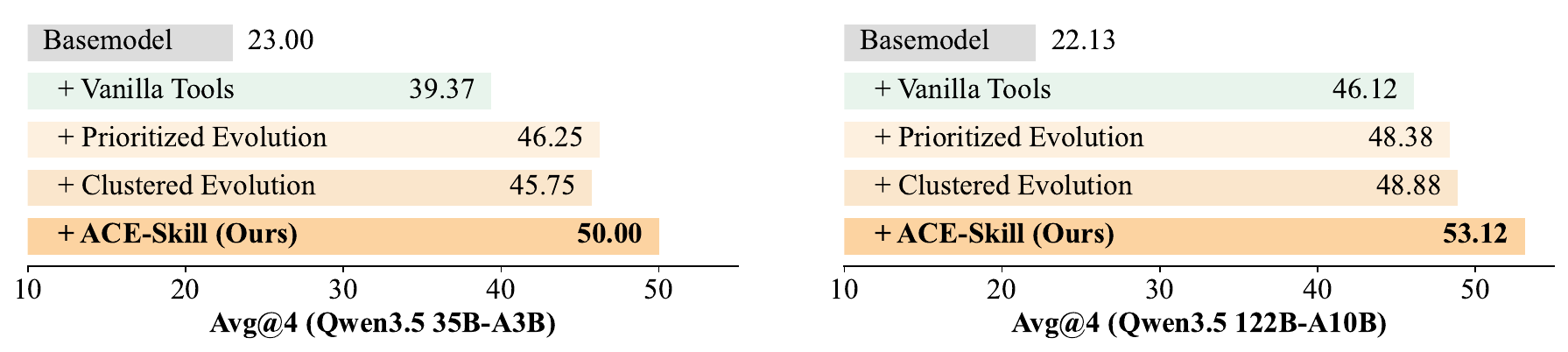}
        \caption{Ablation of proposed modules on TIR-Bench (Avg@4). \textbf{Left}: Qwen3.5-35B-A3B. \textbf{Right}: Qwen3.5-122B-A10B. Each module provides independent gains, and their combination (\ourmethod{}) yields super-additive improvement on both scales.}
        \label{fig:ablation_module}
    \end{figure}

    \noindent\textbf{Ablation of Modules.}
    To disentangle the contributions of the two core modules, we evaluate each module 
    independently on TIR-Bench (Figure~\ref{fig:ablation_module}). 
    The Prioritized Sampler alone lifts the baseline from 39.37 to 46.25 Avg@4. The Clustered Organizer alone reaches 45.75. 
    Their combination (\ourmethod{}, 50.00) yields the highest performance, confirming that the two modules reinforce each other: 
    targeted sampling populates more coherent clusters, and cleaner clusters sharpen the sampler's reward signal. 
    The same trend holds on Qwen3.5-122B-A10B, demonstrating that the synergy is robust across model scales.

    \begin{figure}[t]
        \begin{minipage}[c]{0.48\textwidth}
            \centering
            \captionof{table}{Component-level ablation on TIR-Bench (Avg@4). We evaluate the impact of removing individual design choices from the sampler (base exploration $\epsilon$ and difficulty bias) and the organization module (skill and experience clustering). \textbf{Bold} indicates the best result per column.}
            \label{tab:ablation}
            \renewcommand{\arraystretch}{1.2}
            \setlength{\tabcolsep}{6pt}
            \resizebox{\textwidth}{!}{
            \begin{tabular}{lcc}
            \toprule
            \textbf{Conditions} & \begin{tabular}{c}\textbf{Qwen3.5}\\\textbf{35B-A3B} \end{tabular} & \begin{tabular}{c}\textbf{Qwen3.5}\\\textbf{122B-A10B} \end{tabular} \\ 
            \midrule
            Basemodel & 23.00 & 22.13 \\ 
            \textit{w/} Vanilla Tools & 39.37 & 46.12 \\ 
            \textit{w/} \textbf{\ourmethod{} (Ours)} & \textbf{50.00} & \textbf{53.12} \\ 
            \rowcolor{black!10}\multicolumn{3}{c}{\textit{Ablation of Components}} \\ 
            \textit{w/o} Base Exploration & 48.88 & 52.00 \\
            \textit{w/o} Difficulty Bias & 47.50 & 51.25 \\ 
            \textit{w/o} Skill Cluster & 46.75 & 51.62 \\ 
            \textit{w/o} Experience Cluster & 48.25 & 53.00\\ 
            \bottomrule
            \end{tabular}
            }
        \end{minipage}
        \hfill
        \begin{minipage}[c]{0.48\textwidth}
            \centering
            \includegraphics[width=\textwidth]{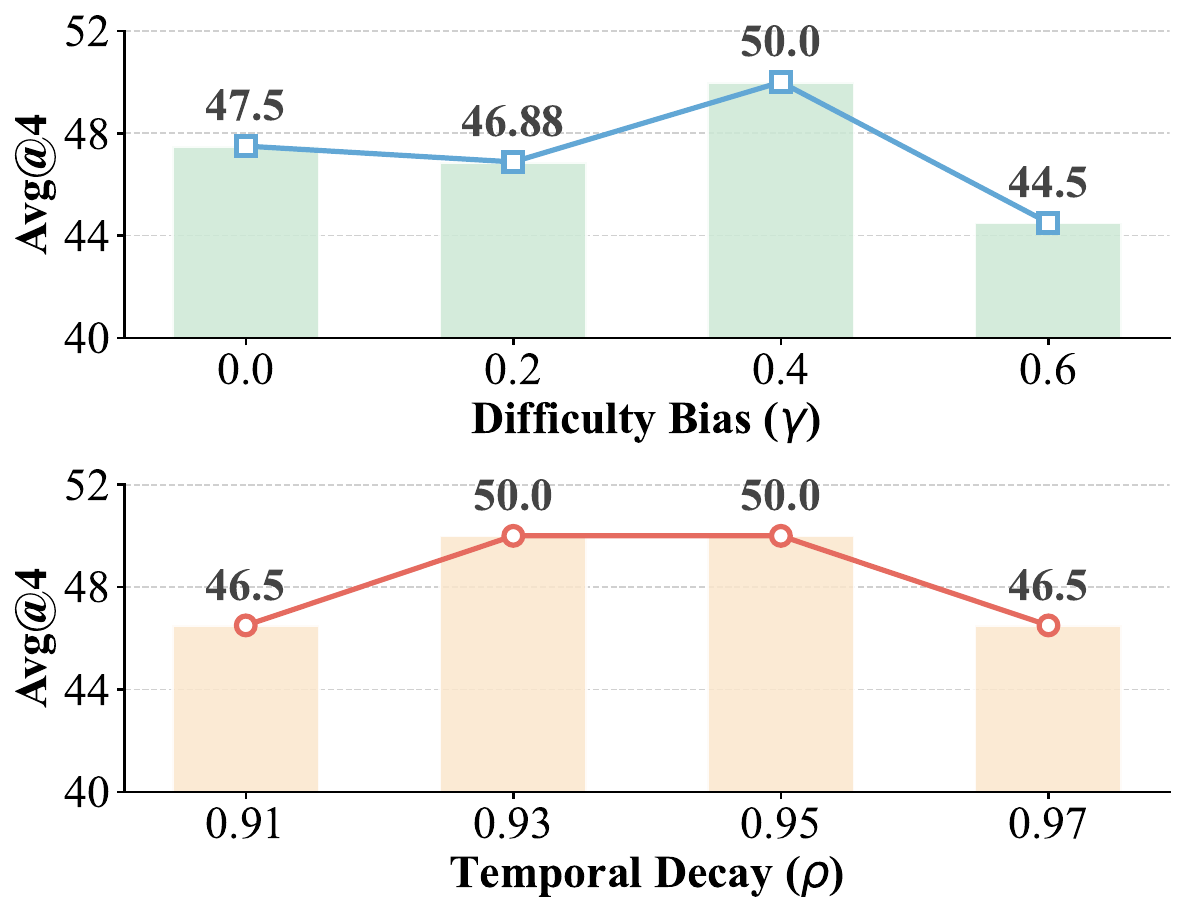}
            \vspace{-0.6cm}
            \captionof{figure}{Hyperparameter sensitivity on TIR-Bench (Avg@4). \textbf{Top}: difficulty bias $\gamma$ ($\rho\!=\!0.95$). \textbf{Bottom}: temporal decay $\rho$ ($\gamma\!=\!0.4$).}
            \label{fig:parameter_analysis}
        \end{minipage}
    \end{figure}
    
    \noindent\textbf{Ablation of Components in Modules.}
    We further ablate individual components within each module. 
    Removing the base exploration term $\epsilon$ (\textit{w/o} epsilon) reduces Avg@4 from 50.00 to 48.88, indicating that a minimum exploration floor is necessary to prevent the sampler from prematurely ignoring under-explored samples. 
    Disabling the difficulty bias (\textit{w/o} difficulty bias) leads to a larger drop (47.50), confirming that up-weighting challenging samples is critical for data utility. 
    On the organization side, removing skill clustering (\textit{w/o} skill cluster, 46.75) hurts more than removing experience clustering (\textit{w/o} experience cluster, 48.25), suggesting that category-level strategic guidance has a broader impact than instance-level tactical lessons when isolation is lost.

    \noindent\textbf{Ablation of Hyperparameter Sensitivity.}
    We analyze the sensitivity of two key hyperparameters: the hardness bias $\gamma$ and the temporal decay $\rho$ on TIR-Bench (Figure~\ref{fig:parameter_analysis}). Performance is best when $\rho$ balances memory retention and forgetting ($\rho \in \{0.93, 0.95\}$), and drops when the decay is too aggressive or too conservative. For $\gamma$, moderate bias toward difficult samples helps, but excessive bias sacrifices coverage and hurts overall performance. Based on these results, we adopt $\rho\!=\!0.95$ and $\gamma\!=\!0.4$ in all other experiments.

    \subsection{Case Study}

    \definecolor{baselinebg}{HTML}{FDF6F6}
    \definecolor{baselineframe}{HTML}{C0392B}
    \definecolor{oursbg}{HTML}{F2FAF2}
    \definecolor{oursframe}{HTML}{27AE60}
    \definecolor{dimgray}{HTML}{888888}
    \definecolor{seccolor}{HTML}{555555}

    \begin{figure*}[t]
        \centering
        \begin{minipage}[t]{\textwidth}
        \begin{tcolorbox}[
            colback=white,
            colframe=black!60,
            coltitle=black,
            fonttitle=\bfseries\scriptsize,
            title={Task: TIR-Bench maze\_369},
            boxrule=0.7pt, arc=2pt,
            left=2pt, right=2pt, top=2pt, bottom=2pt
        ]
        \begin{minipage}[t]{0.20\linewidth}
        \centering
        \vspace{0pt}
        \includegraphics[width=\linewidth]{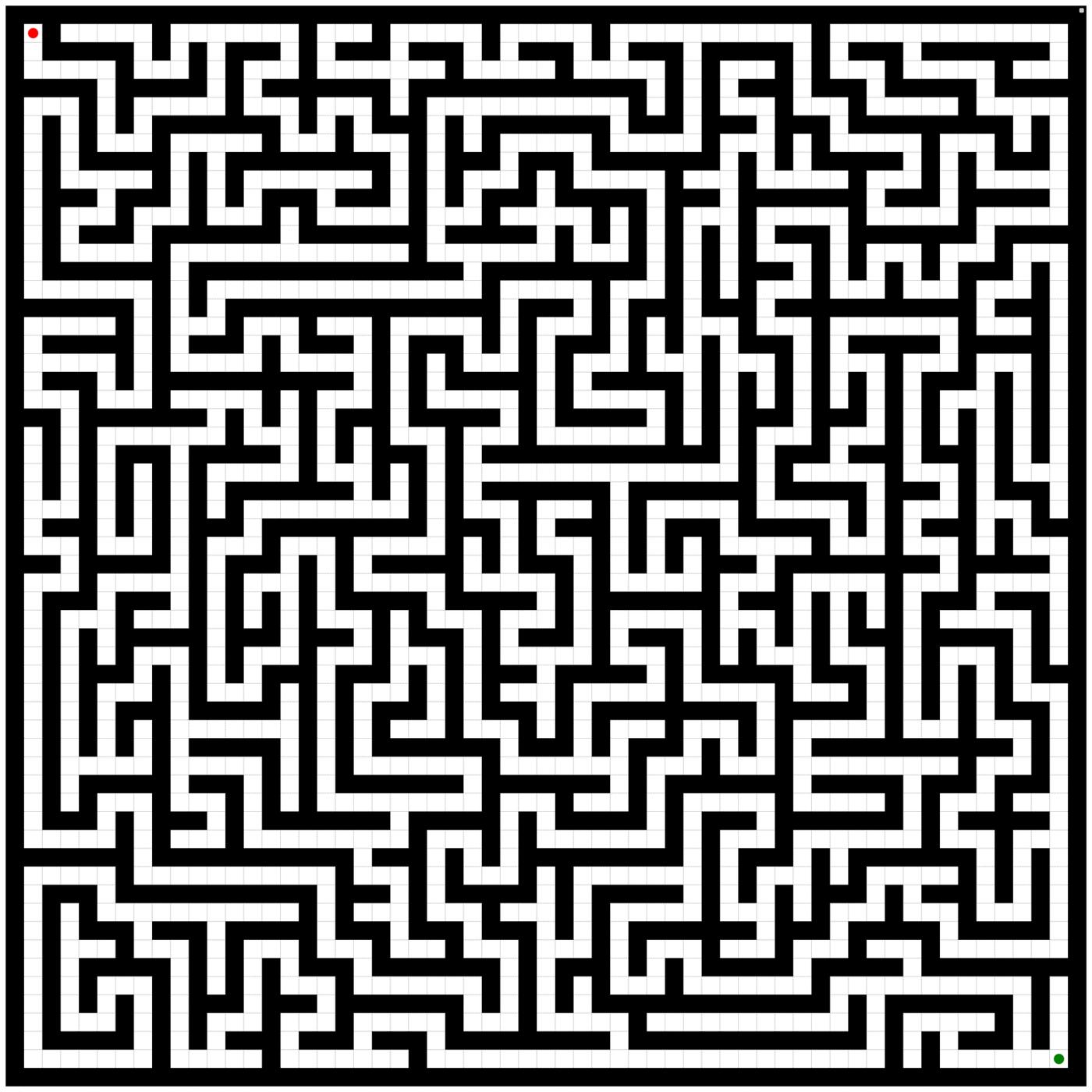}
        \end{minipage}
        \hfill
        \begin{minipage}[t]{0.78\linewidth}
        \vspace{2pt}
        {\fontsize{5.5pt}{6.8pt}\selectfont
        \newcommand{\opt}[1]{\par\hangindent=1.2em\hangafter=1\noindent\textbf{#1}}
        \textbf{Q:} Please complete a maze game shown in the figure. Starting from the \textcolor{red}{\textbf{red ball}} in the top-left corner, navigate the maze to reach the \textcolor{green!60!black}{\textbf{green ball}} in the bottom-right corner. `R' means move one step to the right, `L' means move one step to the left, `U' means move one step up, and `D' means move one step down. Which of the following options can successfully lead out of the maze?\par
        \vspace{2pt}
        \opt{A.} LDUUUUDRDDDRRDRLURDUDLLLRLDDLRDDUDLRUULRLLLRURRUDDLLRLRUURLR...LDRDRDUURDLLDDD \textcolor{dimgray}{(738 chars)}\par
        \vspace{1pt}
        \opt{B.} DLDDRDDRRDDLRDDDLRDRDUDUDRDDDRRRDDDLRLDRRURRDLRRRDRURLDDRRR...RDDRDDRDDURRDDR \textcolor{dimgray}{(180 chars)}\par
        \vspace{1pt}
        \opt{C.} DDRRRRDDDDRRUURRRRUUUURRRRDDLLDDRRDDRRUURRDDRRDDDDLLLLUULLL...UURRRRRRRRDDDDRR \textcolor{dimgray}{(724 chars)}\par
        \vspace{1pt}
        \opt{D.} RURDDRLRRRURDRRRDDDDDDURDULDDDRURRDRRRRLDRRDRURDUDRDRRDRLD...RDLRLDRRRRUDRDDRR \textcolor{dimgray}{(170 chars)}\par
        \vspace{1pt}
        \opt{E.} UDRUURRULLDULRURUURLRDDDULLRRDUULUUUULRLUURRUUDDDULRDRLRRDU...DLLUULDDDLLUDDRR \textcolor{dimgray}{(744 chars)}\par
        \vspace{1pt}
        \opt{F.} No answer\par
        \vspace{1pt}
        \textbf{Ground Truth: \textcolor{green!60!black}{C}}}
        \end{minipage}
        \end{tcolorbox}
        \end{minipage}

        {\centering\scriptsize\textcolor{dimgray}{(Skills, Experiences and Traces below are condensed summaries.)}\par}
        \vspace{2pt}
        \begin{minipage}[t]{0.485\textwidth}
            \begin{tcolorbox}[
                colback=baselinebg,
                colframe=baselineframe,
                coltitle=white,
                fonttitle=\bfseries\scriptsize,
                title={\xmark~w/o Cluster: Activated Skill},
                boxrule=0.7pt, arc=2pt,
                left=3pt, right=3pt, top=2pt, bottom=2pt,
                equal height group=skillrow
            ]
            \begin{Verbatim}[fontsize=\fontsize{5pt}{6.2pt}\selectfont, commandchars=\\\{\}, codes={\catcode`$=3\catcode`^=7}]
    \textcolor{seccolor}{\textbf{GlobalVisualReasoningLibrary}}
    \textcolor{seccolor}{\textbf{## 1. Input \& Preprocessing}}
    - Variable \& type integrity; normalize to NumPy array.
    - Orientation correction; CLAHE contrast enhancement.
    - Segmentation to binary masks; ROI bounding boxes.
    \textcolor{seccolor}{\textbf{## 2. Core Primitives}} (code templates)
    - load\_and\_convert(); enhance\_contrast(); find\_optimal\_grid()
    \textcolor{seccolor}{\textbf{## 3. Task Workflows}}
    - 4.1 Information Extraction (OCR / symbols)
    - 4.2 Quantitative Estimation (counting / metrics)
    - \textbf{4.3 Spatial Logic (grids / paths)}
    - 4.4 Metric Extraction (instruments / dials)
    \textcolor{seccolor}{\textbf{## 4. Validation \& Robustness}}
    - Overlay requirement; visual truth priority.
    - Common pitfalls: ROI limits, scale ambiguity, box vs object.
    \textcolor{baselineframe}{(Unified SOP covering all visual tasks; maze-solving is one of}
    \textcolor{baselineframe}{ four workflows — generic but not specialized for grids.)}
            \end{Verbatim}
            \end{tcolorbox}
            \end{minipage}
            \hfill
            \begin{minipage}[t]{0.485\textwidth}
            \begin{tcolorbox}[
                colback=oursbg,
                colframe=oursframe,
                coltitle=white,
                fonttitle=\bfseries\scriptsize,
                title={\cmark~w/ Cluster: Activated Skill},
                boxrule=0.7pt, arc=2pt,
                left=3pt, right=3pt, top=2pt, bottom=2pt,
                equal height group=skillrow
            ]
            \begin{Verbatim}[fontsize=\fontsize{5pt}{6.2pt}\selectfont, commandchars=\\\{\}, codes={\catcode`$=3\catcode`^=7}]
    \textcolor{seccolor}{\textbf{VisualGridSolver}}
    \textcolor{seccolor}{\textbf{## 1. Perception (Visual $\to$ Logical)}}
    - Grayscale + adaptive threshold; crop to ROI bounding box.
    - Dimension: run-length transitions $\to$ median cell size $\to$ N.
      Fallback: iterate adjacent sizes ($\pm$1 to $\pm$5).
    - HSV marker detection; centroids $\to$ grid indices.
    - Force start/end cells walkable (occlusion override).
    \textcolor{seccolor}{\textbf{## 2. Reasoning (Logical $\to$ Decision)}}
    - Heuristic pruning: discard options with invalid first moves.
    - BFS connectivity check before full simulation.
    - Path simulation: step-by-step; abort on wall/OOB collision.
    - Fallback: compare BFS shortest path via edit distance.
    \textcolor{seccolor}{\textbf{## Code Template:}} extract\_grid\_structure()+
      simulate\_path() \textcolor{oursframe}{(full pipeline; truncated)}
    \textcolor{seccolor}{\textbf{## Failure Modes}} (7 categories w/ resolutions)
    \textcolor{oursframe}{(Specialized for grid/maze tasks; includes executable code,}
    \textcolor{oursframe}{ iterative dimension search, and BFS pre-validation.)}
            \end{Verbatim}
            \end{tcolorbox}
            \end{minipage}    
        \vspace{4pt}
        
        \begin{minipage}[t]{0.485\textwidth}
        \begin{tcolorbox}[
            colback=baselinebg,
            colframe=baselineframe,
            coltitle=white,
            fonttitle=\bfseries\scriptsize,
            title={\xmark~w/o Cluster: Retrieved Experiences},
            boxrule=0.7pt, arc=2pt,
            left=3pt, right=3pt, top=2pt, bottom=2pt,
            equal height group=exprow
        ]
        \begin{Verbatim}[fontsize=\fontsize{5pt}{6.2pt}\selectfont, commandchars=\\\{\}, codes={\catcode`$=3\catcode`^=7}]
\textcolor{baselineframe}{[E1]} Isolate inner playable area; largest connected component of non-border pixels.
\textcolor{baselineframe}{[E2]} Otsu/morphology; color masking; fill-factor 0.5--0.9; BFS if fails.
  \textcolor{baselineframe}{↑ first half is generic CV preprocessing — not maze-specific}
\textcolor{baselineframe}{[E3]} Grid dims from pixel runs; separate row/col stats; median cell size.
\textcolor{baselineframe}{[E4]} Local visual validation; discard wall contradictions without simulation.
\textcolor{baselineframe}{[E5]} Compare path length \& turn sequence; distinguish minor vs fundamental mismatch.
\textcolor{baselineframe}{[E6]} Longest valid prefix; tolerate minor discrepancies; avoid premature No Answer.
\textcolor{baselineframe}{[E7]} Early termination: abort simulation on wall/boundary collision.
\textcolor{baselineframe}{[E8]} When transcribing sequences of similar symbols (e.g., digits), compare
     ambiguous characters against clear instances. Verify structural consistency.
  \textcolor{baselineframe}{↑ symbol-transcription tip — irrelevant to maze solving}
\textcolor{baselineframe}{(E2 is half-noisy; E8 is cross-type noise.)}
        \end{Verbatim}
        \end{tcolorbox}
        \end{minipage}
        \hfill
        \begin{minipage}[t]{0.485\textwidth}
        \begin{tcolorbox}[
            colback=oursbg,
            colframe=oursframe,
            coltitle=white,
            fonttitle=\bfseries\scriptsize,
            title={\cmark~w/ Cluster: Retrieved Experiences},
            boxrule=0.7pt, arc=2pt,
            left=3pt, right=3pt, top=2pt, bottom=2pt,
            equal height group=exprow
        ]
        \begin{Verbatim}[fontsize=\fontsize{5pt}{6.2pt}\selectfont, commandchars=\\\{\}, codes={\catcode`$=3\catcode`^=7}]
\textcolor{oursframe}{[E1]} Automate grid-dimension scoring; iterate candidate sizes from coarse
     to fine; pick the resolution that maximizes cell-boundary alignment.
\textcolor{oursframe}{[E2]} Verify code-interpreter variable names against system specs; fix
     NameErrors immediately instead of guessing.
\textcolor{oursframe}{[E3]} Verify input image data type (PIL vs array) immediately; resolve
     type mismatches early to prevent downstream errors.
\textcolor{oursframe}{[E4]} Digitize grids via center sampling; HSV blob detection + centroids;
     morphological dilation to connect fragments.
\textcolor{oursframe}{[E5]} If pathfinding fails, suspect grid extraction errors. Adjust dims
     $\pm$1 to $\pm$5. Prioritize BFS connectivity before comparing options.
\textcolor{oursframe}{(Covers grid extraction, dimension scoring, pathfinding fallback,}
\textcolor{oursframe}{ type safety, and code-interpreter debugging.}
        \end{Verbatim}
        \end{tcolorbox}
        \end{minipage}

        \vspace{4pt}

        \begin{minipage}[t]{0.485\textwidth}
        \begin{tcolorbox}[
            colback=baselinebg,
            colframe=baselineframe,
            coltitle=white,
            fonttitle=\bfseries\scriptsize,
            title={\xmark~w/o Cluster: Inference Trace},
            boxrule=0.7pt, arc=2pt,
            left=3pt, right=3pt, top=2pt, bottom=2pt
        ]
        \begin{Verbatim}[fontsize=\fontsize{5pt}{6.2pt}\selectfont, commandchars=\\\{\}, codes={\catcode`$=3\catcode`^=7}]
\textcolor{seccolor}{\textbf{Turn 0:}} HSV finds Red=(36,39), Green=(1351,1355).
  Median run=1.0 (noise). SyntaxError crashes code.
  \textcolor{baselineframe}{$\to$ Markers found but grid build fails.}
\textcolor{seccolor}{\textbf{Turn 1--2:}} Otsu+GaussianBlur; ROI 1386$\times$1386.
  N=58, Start=(1,1), End=(55,56). BFS len=301.
  \textcolor{baselineframe}{$\to$ End off by one; path inconsistent with options.}
\textcolor{seccolor}{\textbf{Turn 3--4:}} Fix bbox $\to$ N=58, BFS=None (TypeError).
  Search N=55--69 $\to$ picks N=55; BFS=None.
  \textcolor{baselineframe}{$\to$ Start/end cells land on walls (gray$\approx$76).}
\textcolor{baselineframe}{  (...turns 5--8: N=55 w/ forced start/end; all ``No solution'')}
\textcolor{seccolor}{\textbf{Turn 9:}} Search N=50--69 $\to$ picks N=64.
  morph\_open produces fake all-R-then-D path; no option matches.
  \textcolor{baselineframe}{$\to$ 10 turns; never found correct N=59.}
\textcolor{baselineframe}{Pred: F (No answer) $\neq$ GT: C \xmark}
        \end{Verbatim}
        \end{tcolorbox}
        \end{minipage}
        \hfill
        \begin{minipage}[t]{0.485\textwidth}
        \begin{tcolorbox}[
            colback=oursbg,
            colframe=oursframe,
            coltitle=white,
            fonttitle=\bfseries\scriptsize,
            title={\cmark~w/ Cluster: Inference Trace},
            boxrule=0.7pt, arc=2pt,
            left=3pt, right=3pt, top=2pt, bottom=2pt
        ]
        \begin{Verbatim}[fontsize=\fontsize{5pt}{6.2pt}\selectfont, commandchars=\\\{\}, codes={\catcode`$=3\catcode`^=7}]
\textcolor{seccolor}{\textbf{Turn 0:}} np.array(original\_image). HSV inRange
  for red/green; grayscale thresh 128; median cell=24.
  N=58 initial estimate. Start$\approx$(42,42).
  \textcolor{oursframe}{$\to$ HSV marker detection succeeds on first try.}
\textcolor{seccolor}{\textbf{Turn 1--2:}} Cell-variance scoring over N=35--61.
  N=35 too coarse; N=61 end misaligned.
  \textcolor{oursframe}{$\to$ Iterative sizing guided by skill (E1).}
\textcolor{seccolor}{\textbf{Turn 3:}} Variance search N=55--69 $\to$ best N=59.
  Start=(1,1), End=(57,57). Simulate A--E:
  only \textcolor{oursframe}{C reaches (57,57) in 724 steps. $\checkmark$}
  BFS shortest path = 724, consistent with C.
  \textcolor{oursframe}{$\to$ Converged in 4 turns; BFS validates C.}
\textcolor{seccolor}{\textbf{Turn 4:}} Final answer output.
\textcolor{oursframe}{Pred: C = GT: C \cmark}
        \end{Verbatim}
        \end{tcolorbox}
        \end{minipage}

        \caption{\textbf{Case Study: Clustered vs.\ Non-Clustered Pipeline on a Maze Task.} From top to bottom: task, activated skills, retrieved experiences, and inference traces.}
        \label{fig:case_study_evo}
    \end{figure*}
    
    To intuitively illustrate the qualitative impact of Clustered Evolution Organization, we compare the skill documents evolved \emph{with} and \emph{without} cluster isolation on the same visual grid-solving task from TIR-Bench. As shown in Figure~\ref{fig:case_study_evo}, the impact of clustered evolution propagates through the entire inference pipeline.
    
    The two systems produce different skill documents: the cluster-specialized \texttt{VisualGridSolver} features a layered Perception$\to$Reasoning pipeline, an end-to-end executable code template with run-length-based dimension estimation, occlusion-override rules, and a structured 7-category failure table, whereas the monolithic \texttt{GlobalVisualReasoningLibrary} covers four broad workflow categories (OCR, counting, spatial logic, metric extraction) with generic validation heuristics but no executable code or grid-specific recovery protocols.
    
    For experience retrieval, the non-clustered system draws eight experiences from the global pool. While most are relevant to grid extraction and path simulation, the retrieved set is diluted: one experience is only partially useful, with its first half devoted to generic CV preprocessing rather than maze-specific logic, and another is pure cross-type noise, namely a symbol-transcription tip entirely irrelevant to maze solving. In contrast, the clustered system retrieves five tightly focused experiences from the grid-solving cluster, covering automated dimension scoring, input variable verification, data type checking, grid digitization with HSV-based overlay validation, and BFS connectivity with iterative dimension adjustment.
    
    During inference, this gap becomes decisive. The baseline agent detects the red and green markers via HSV in Turn~0 but immediately crashes on a \texttt{SyntaxError}; subsequent turns estimate $N\!=\!58$ and search over $N\!=\!55$ to $69$ yet repeatedly place start/end cells on walls (gray$\approx$76), and after 10 turns it settles on $N\!=\!64$ with a spurious all-R-then-D path, concluding ``No answer''. The cluster-guided agent, prompted by its skill's variable-verification and HSV-fallback strategies, successfully detects both markers on the first try, applies cell-variance scoring to iteratively refine the grid dimension from $N\!=\!58$ to $N\!=\!59$, and simulates all options in Turn~3, finding that only C reaches the endpoint (57,57) in 724 steps with BFS confirmation, answering~C within 4 turns.

\section{Related Work} \label{sec:related_work}
    \noindent\textbf{Self-Evolved Agents via Experience and Skill Bootstrapping.}
    A growing body of work enables LLM agents to self-improve by distilling reusable knowledge from execution trajectories. 
    Reflexion~\citep{shinn2023reflexion} pioneers verbal reinforcement by storing textual reflections for future reasoning. 
    Subsequent methods extend this idea to richer knowledge representations: 
    XSkill~\citep{jiang2026xskill} introduces a dual-stream framework that jointly accumulates experiences and skills through visually grounded summarization, while Memento-Skills~\citep{zhou2026memento} treats skill files as persistent memory that evolves via read-write reflective learning. 
    Trace2Skill~\citep{ni2026trace2skill} distills trajectory-local lessons into transferable skill documents through hierarchical consolidation, and SkillX~\citep{wang2026skillx} constructs multi-level skill hierarchies with iterative refinement and exploratory expansion. Other lines explore collective evolution across users~\citep{ma2026skillclaw}, multimodal generation with agent memory~\citep{he2026gems}, learnable memory skills~\citep{zhang2026memskill}, prompt-level self-improvement at scale~\citep{li2026combee}, and exploration-based skill discovery~\citep{yang2025exif}. 
    Despite their diversity, these methods typically aggregate all knowledge into a single global pool, incurring cross-type retrieval noise as sample diversity grows, and most operate at a single granularity (either experiences or skills). 
    \ourmethod{} addresses both gaps by organizing knowledge into cluster-isolated banks, each maintaining dual-granularity stores (tactical experiences and strategic skills), and couples this organization with a prioritized sampler that actively targets informative samples.
    
    \noindent\textbf{Skill Representation, Evaluation, and Lifecycle.}
    The recent survey SoK~\citep{jiang2026sok} formalizes the full skill lifecycle spanning discovery, distillation, storage, composition, evaluation, and update, and identifies seven design patterns for skill-based agent systems. 
    SkillsBench~\citep{li2026skillsbench} provides the first systematic benchmark of skill utility, revealing that curated skills boost pass rates by 16.2pp while self-generated skills yield no average gain, underscoring the importance of quality control. 
    On the representation front, ContractSkill~\citep{lu2026contractskill} makes skills explicit and repairable through executable contracts, and SkillReducer~\citep{gao2026skillreducer} demonstrates a less-is-more effect where compressing non-actionable content reduces context dilution. 
    PolySkill~\citep{yu2026polyskill} decouples a skill's abstract goal from its concrete implementation for cross-site generalization, while CoEvoSkills~\citep{zhang2026coevoskills} co-evolves skill generation and surrogate verification without ground-truth supervision. 
    As skill libraries scale, SkillRouter~\citep{zheng2026skillrouter} addresses the routing problem with a compact retrieve-and-rerank pipeline for large-scale registries. 
    ASDA~\citep{yim2026asda} proposes training-free skill distillation through error-corrective learning for domain adaptation. 
    In contrast to methods that rely on explicit code-based skill formats or external routers, \ourmethod{} represents skills as structured natural-language documents within semantically coherent clusters, naturally resolving routing through cluster assignment while maintaining compactness via word-budget refinement.
    
    \noindent\textbf{Reinforcement Learning for Agent Skill Acquisition.}
    A separate line of research focuses on internalizing skills into model parameters through modern reinforcement learning strategies \citep{chu2026gpg,ji2026tree,wang2026visually,dai2026harder}. 
    D2Skill~\citep{tu2026dynamic} jointly trains the policy and a dual-granularity skill bank using hindsight utility signals, while SKILL0~\citep{lu2026skill0} progressively withdraws skill context during RL training to achieve zero-shot internalization. 
    SkillRL~\citep{xia2026skillrl} co-evolves a hierarchical skill library and the agent policy through recursive reinforcement learning, and ARISE~\citep{li2026arise} employs hierarchical RL with a tiered skill library where a manager selects skills to condition worker rollouts. 
    SAGE~\citep{wang2026sage} augments GRPO with sequential rollouts that accumulate skills along task chains. Beyond skill-centric methods, DGO~\citep{bai2026dgo} constructs an experience bank to jointly guide exploration and internalization during RLVR training, 
    MetaClaw~\citep{xia2026metaclaw} combines continual meta-learning with opportunistic LoRA fine-tuning, and LEAFE~\citep{ge2026leafe} distills reflective experience-guided corrections via supervised fine-tuning. 
    While these RL-based approaches can deeply embed skills into model weights, they require substantial computational overhead for gradient updates and produce model-specific knowledge that does not transfer across architectures. 
    \ourmethod{} takes a complementary, training-free approach: it enhances agent capabilities purely through prompt augmentation, making the accumulated knowledge lightweight, model-agnostic, and directly transferable across different model scales without retraining.

\section{Limitation and Discussion}
    While \ourmethod{} demonstrates substantial improvements in multimodal tool-use capabilities and successfully mitigates data inefficiency and knowledge interference, we identify a few areas for broader discussion and future exploration.

    \begin{itemize}[leftmargin=*, itemsep=1pt, topsep=2pt]
        \item \textbf{Computational Cost in Offline Bootstrapping.} The self-evolution process, particularly the synthesis of strategic skills and tactical experiences, introduces a moderate computational overhead during the offline training phase. However, we emphasize that this is a one-time cost. Thanks to our \emph{Prioritized Evolution Sampler}, the overall sample complexity and training time are already significantly reduced compared to standard uniform sampling baselines. Furthermore, once the agent is trained, the online inference cost remains identical to or even lower than the base model due to more accurate, zero-shot tool invocations. Future work could explore using lighter models as the robust verifier to further democratize the bootstrapping process.
    
        \item \textbf{Scalability of the Tool Library.} In this work, we focus on four core, representative tool categories (Code Interpreter, Web Search, Image Search, and Visit) that effectively cover a vast majority of scenarios in existing multimodal benchmarks. Real-world applications, however, may involve an open-ended, long-tail distribution of specialized APIs. Fortunately, the \emph{Clustered Evolution Organization} is inherently modular. Adapting \ourmethod{} to an expanding tool library simply involves initializing new semantic clusters, allowing the agent to continually acquire new skills without suffering from catastrophic forgetting of previously learned capabilities.
    
        \item \textbf{Dependence on Base Model Capacity.} \ourmethod{} successfully elevates open-source models (\eg, Qwen3.5-35B) to match or surpass proprietary systems. Naturally, the absolute upper bound of the learned skills is partially influenced by the base model's fundamental reasoning and multimodal understanding capabilities. To mitigate the reliance on massive backbones for practical deployment, our experiments have already demonstrated strong zero-shot transferability: knowledge bootstrapped by a larger model can be directly applied to smaller-scale models (\eg, 9B and 4B). This cross-scale distillation presents a highly promising pathway for deploying highly capable tool-use agents on edge or resource-constrained devices.
    \end{itemize}

\section{Conclusion} \label{sec:conclusion}
    In this paper, we identify two critical, entangled bottlenecks in current self-evolution paradigms for multimodal agents: the poor data utility of sequential sample traversal and the knowledge interference caused by cross-type noise in a global experience pool. To address these, we propose \ourmethod{}, a co-evolutionary framework that jointly optimizes data allocation and knowledge organization. \ourmethod{} replaces standard uniform sampling with a \emph{Prioritized Evolution Sampler} guided by a Bayesian value tracker, and restructures the global experience pool into a \emph{Clustered Evolution Organization} that strictly isolates dual-granularity knowledge (tactical experiences and strategic skills). Empirical evaluations demonstrate that this tightly coupled design consistently outperforms proprietary and open-source methods on general tool-use benchmarks and exhibits strong zero-shot transferability. Ultimately, by dismantling the vicious cycle between poor data utilization and noisy retrieval, we hope this co-evolutionary paradigm offers practical insights and a scalable recipe for building self-improving multimodal agents.

\bibliography{main}
\bibliographystyle{unsrtnat}

\end{document}